\documentclass{article}
\usepackage{spconf,amsmath,graphicx}

\usepackage{hyperref}
\usepackage{url}
\usepackage{amssymb}

\usepackage{caption}
\usepackage{subcaption}
\usepackage{cite}
\usepackage{booktabs}
\usepackage{algorithmic}
\usepackage{algorithm}
\usepackage{verbatim}

    \title{Semi-Recurrent CNN-based VAE-GAN for Sequential Data Generation}

\name{Mohammad Akbari and Jie Liang}
\address{School of Engineering Science, Simon Fraser University, Burnaby, BC, Canada}

\begin{document}
%\ninept
%
\maketitle
\begin{abstract}
A semi-recurrent hybrid VAE-GAN model for generating sequential data is introduced. In order to consider the spatial correlation of the data in each frame of the generated sequence, CNNs are utilized in the encoder, generator, and discriminator. The subsequent frames are sampled from the latent distributions obtained by encoding the previous frames. As a result, the dependencies between the frames are maintained. Two testing frameworks for synthesizing a sequence with any number of frames are also proposed. The promising experimental results on piano music generation indicates the potential of the proposed framework in modelling other sequential data such as video.
\end{abstract}
\section{Introduction}
\label{Introduction}

% introduction
One important problem in unsupervised learning is generating sequential data such as music. Recurrent Neural Networks (RNNs) and Long Short Term Memory Networks (LSTMs) have shown considerable performance in this area. However, they have difficulties in handling the vanishing and the exploding gradient problems \cite{pascanu2013difficulty}. In order to deal with these issues, RNNs have been combined with the most recent deep generative architectures such as Variational Auto-encoders (VAEs) and Generative Adversarial Networks (GANs) \cite{fabius2014variational, tikhonov2017music, hadjeres2017glsr, mogren2016c, guimaraes2017objective, yu2017seqgan}, which are typically used for learning complex structures of data.

VAEs are generally easy to train, but the generated results have low quality due to imperfect measures such as the squared error. On the other hand, GANs generate samples with higher quality, but they suffer from training instability. In order to improve the training process and the quality of the generated samples, some researchers suggested hybrid VAE-GAN models \cite{larsen2015autoencoding, makhzani2015adversarial}.

Although most of the sequential data generation methods are based on RNNs, some recent works have shown that Convolutional Neural Networks (CNNs) are also capable of generating realistic sequential data such as music \cite{oord2016wavenet, yang2017midinet}. One advantage of CNNs is that they are practically faster to train and easier to parallelize than RNNs. In addition, applying convolutions to the time dimension can result in significant performance in some applications \cite{briot2017deep}.

\begin{figure}
\begin{minipage}[b]{0.4\linewidth}
 \centering
  \centerline{\includegraphics[width=3cm]{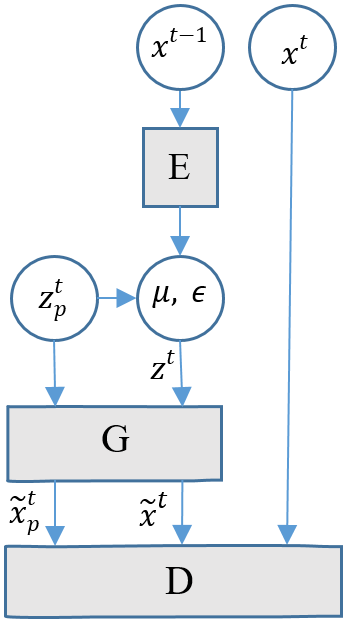}}
  %\centerline{Subtraction}\medskip
  \subcaption{Training framework.}
  \label{fig:trainingframework}
\end{minipage}
\begin{minipage}[b]{0.65\linewidth}
 \centering
  \centerline{\includegraphics[width=4cm]{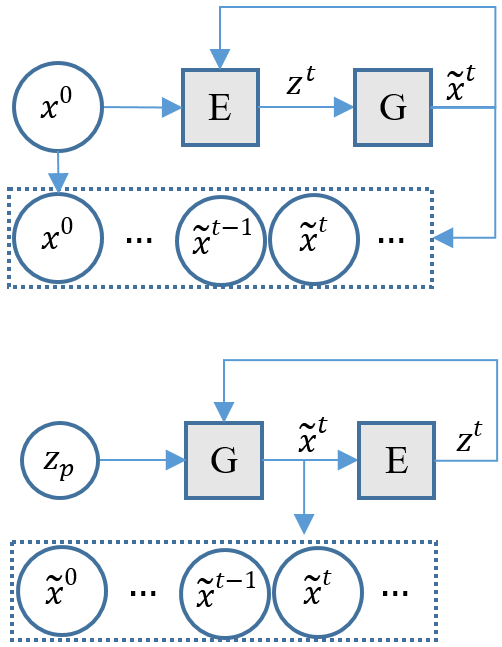}}
  \subcaption{Two testing frameworks.}
  \label{fig:testingframework}
\end{minipage}
\caption{The training and testing frameworks of the proposed semi-recurrent hybrid VAE-GAN model ($E$: encoder, $G$: generator, and $D$: discriminator).}
\label{fig:framework}
\end{figure}

Considering the sequential data generation as a problem of generating a sequence of discrete frames, two problems need to be addressed: strong spatial correlation of the data in each of the frames, and the dependencies between them (temporal correlation). In this work, we propose a semi-recurrent convolution-based VAE-GAN for generating a sequence of individual frames where the above problems are efficiently addressed. In order to maintain strong local correlation of the data in each frame generated, we use CNN, which is a very effective architecture for this matter. Moreover, each frame is generated from the latent distribution of the previous frame encoded by an encoder. As a result, the dependencies across the frames are also kept. 

Figure \ref{fig:framework} illustrates the overall training and testing frameworks proposed in this work. The model includes an encoder, a generator (decoder), and a discriminator. To the best of our knowledge, this is the first hybrid VAE-GAN framework introduced for generating sequential data. The feasibility of this model is evaluated on piano music generation, which shows that the proposed framework is a viable way of training networks that model music, and has potential for modelling many other types of sequential data such as videos.

%The paper is organized as follows. In Section \ref{Related Works}, a brief background of the deep generative VAE and GAN models is given, which is followed by the works related to generating sequential data, especially music. The proposed approach is then formulated and described in Section \ref{Proposed Method: Hybrid VAE-GAN}. In Section \ref{Experiments}, the feasibility of this model is finally investigated and evaluated on piano music generation. We conclude that the proposed framework is a viable way of training networks that model music, and see potential for also modelling many other types of sequential continuous data.

%he discrete nature of the symbolic representation

\section{Preliminaries and Related Works}
\label{Related Works}

%Unlike discriminative models that infer outputs based on inputs, generative models generate both inputs and outputs, typically given some hidden parameters. 

%A variety of generative models have been proposed in the past years. Principle Component Analysis (PCA) \cite{turk1991face}, Independent Component Analysis (ICA) \cite{hyvarinen2004independent}, and Gaussian Mixture Model (GMM) \cite{permuter2003gaussian} are considered as the early research of the generative models, which have problems with modeling complex data distributions such as images. Hidden Markov Model (HMM) \cite{starner1997real}, Markov Random Field (MRF) \cite{mnih2010generating}, restricted Boltzmann machines (RBMs) \cite{hinton2006reducing, salakhutdinov2009deep} are the later works, which do not generally learn effective feature representations.

In recent years, deep generative models have achieved significant success, especially in generating natural images \cite{kingma2013auto, goodfellow2014generative, rezende2014stochastic, radford2015unsupervised, salimans2016improved}. In these models, complex structures of the data can be learned using deep architectures with multiple layers. VAEs \cite{kingma2013auto, rezende2014stochastic} and GANs \cite{goodfellow2014generative, radford2015unsupervised, salimans2016improved} are two powerful frameworks for learning deep generative models in an unsupervised manner.

\subsection{Variational Auto-encoder (VAE)}
\label{Variational Auto-encoder}

A VAE consists of an encoder and a decoder \cite{kingma2013auto}. The encoder, denoted by $q(z|x)$, encodes a data sample $x$ to a latent (hidden) representation $z$: $z \sim q(z|x)$. The decoder, denoted by $p(x|z)$, decodes the latent representation back to the probability distribution of the data (in data space): $\hat{x} \sim p(x|z)$.

The VAE regularizes the encoder by imposing a prior over the latent distribution $p(z)$ where $z \sim \mathcal{N}(0, I)$. The loss function of the VAE is the expected log likelihood with a regularizer:
\begin{equation}
\mathcal{L}_{VAE}=- \mathbb{E}_{q(z|x)}[\log p(x|z)]+KL(q(z|x)\|p(z))
\end{equation}
where the first and second terms are the reconstruction loss and a prior regularization term that is the Kullback-Leibler (KL) divergence, respectively.

%Although VAEs are generally easy to train, the generated results have low quality (e.g., blurry images) due to minimizing the MSE-based reconstruction error.
%VAE [12, 31] pairs a differentiable encoder network with a decoder/generative network. A disadvantage of VAE is that, because of the injected noise and imperfect element-wise measures such as the squared error, the generated samples are often blurry.
%Blurry result due to minimizing the MSE based reconstruction error

\subsection{Generative Adversarial Network (GAN)}
\label{Generative Adversarial Network}

Another popular generative model is GAN in which two models are trained at the same time \cite{goodfellow2014generative}. The generator model $G(z)$ captures the data distribution by mapping the latent $z$ to data space, while the discriminator model $D(x) \in [0, 1]$ estimates the probability that $x$ is a real training sample or a fake sample synthesized by $G$. These two models compete in a two-player minimax game in which the objective function is to find a binary classifier $D$ that discriminates the real data from the fake (generated) ones, and simultaneously encourage $G$ to fit the true data distribution. This goal is achieved by minimizing/maximizing the binary cross entropy: 
\begin{equation}
\mathcal{L}_{GAN} =\mathbb{E}_{x\sim p_{data}(x)}[\log D(x)] + \mathbb{E}_{z\sim p_{z}(z)}[\log(1 - D(G(z)))]
\end{equation}
where $G$ tries to minimize this objective against $D$ that tries to maximize it.

%We train D to maximize the probability of assigning the correct label to both training examples and samples from G

% Generator tries the best to cheat the discriminator by generating more realistic images. Discriminator tries the best to distinguish whether the image is generated by computers or not.

% : the generator network Gen(z) maps latents z to data space while the discriminator network assigns probability y = Dis(x) ∈ [0, 1] that x is an actual training sample and probability 1 − y that x is generated by our model through x = Gen(z) with z ∼ p(z).

% a generative model G that captures the data distribution, and a discriminative model D that estimates the probability that a sample came from the training data rather than G. The training procedure for G is to maximize the probability of D making a mistake. This framework corresponds to a minimax two-player game.

%The generator G implicitly defines a probability distribution pg as the distribution of the samples G(z) obtained when z ∼ pz.

% The GAN objective is to find the binary classifier that gives the best possible discrimination between true and generated data and simultaneously encouraging Gen to fit the true data distribution. We thus aim to maximize/minimize the binary cross entropy:

Although GANs are powerful generative models, they suffer from training instability and low-quality generated samples. Different approaches have been proposed to improve GANs. For example, Wasserstein GAN (WGAN) \cite{arjovsky2017wasserstein} used Wasserstein distance as an objective for training GANs to improve the stability of learning, Laplacian GAN (LAPGANs) \cite{denton2015deep} achieved coarse-to-fine conditional generation through Laplacian pyramids, and Deep Convolutional GAN (DCGAN) \cite{radford2015unsupervised} proposed an effective and stable architecture for $D$ and $G$ using deeper CNNs to achieve remarkable image synthesis results. 
%
%conditional VAE/GAN: CVAE , CGAN (not now)
% 
%hybrid VAE/GAN

%In order to take the advantages of both VAE and GAN into account, hybrid VAE-GAN \cite{larsen2015autoencoding, makhzani2015adversarial} models were proposed in which the training process and the quality of the images generated are improved. In these models the VAE decoder and GAN generator are collapsed into one.
%In order to improve the quality of the images generated by VAEs, hybrid VAE-GAN models were proposed in which GAN training is also improved.
%% VAEs have been successfully used to improve GAN training

\subsection{Sequential Data Generation: Music Generation}
\label{Music Generation}

Different learning-based approaches for sequential data generation, especially music, have been introduced by various researchers. In \cite{eck2002first}, a RNN-based architecture using LSTMs was proposed in which a piano-roll sequence of notes and chords were generated using an iterative feed-forward strategy. %Some limitation of this work includes handcrafted training set and deterministic generation. 
%A decoder feed-forward strategy on a stacked auto-encoder architecture was presented in \cite{sarroff2014musical} to generate music spectrograms. 
%In some works, sampling strategy was used for music generation\cite{boulanger2012modeling, hadjeres2016deepbach}. 
In \cite{boulanger2012modeling} a Restricted Boltzmann Machine (RBM) was utilized for modeling and generating polyphonic music by learning a model from an audio corpus. DeepBach architecture \cite{hadjeres2016deepbach}, which was specialized for Bach's chorales, combined two LSTMs and two feed-forward networks (forward and backward in time networks). 
%BachBot \cite{} is another work for polyphonic music in the style of Bach's chorales.

VAE, as one of the effective approaches considered for content generation, has been used by some researchers in order to generate musical content. In \cite{fabius2014variational}, a VAE-based method named Variational Recurrent Auto-Encoder (VRAE) was proposed in which the encoder and decoder parts were LSTMs. 
%In this work, the encoder sequentially reads each symbol of an input sequence and the decoder synthesizes the output sequence by predicting the next symbol. 
Variational Recurrent Autoencoder Supported by History (VRASH) \cite{tikhonov2017music} used the same architecture as in VRAE, but also used the output of the decoder back into the decoder. 
%As a result, the previous note generated was considered as an added information to the next note. 
In \cite{hadjeres2017glsr}, the objective function used in DeepBach was reformulated using VAE to have a better control on the embedding of the data into the latent space.

Although RNNs are more commonly used to model time-series signals, some non-RNN approaches have been introduced using CNNs \cite{oord2016wavenet, lattner2016imposing, yang2017midinet}. A system for generating raw audio music waveforms named WaveNet was proposed in \cite{oord2016wavenet} in which an extended CNN called dilated causal convolution was incorporated. In this work, the authors argued that dilated convolutions allowed the receptive field to grow longer in a much cheaper way than using LSTMs. Another CNN-based architecture is convolutional RBM (C-RBM) \cite{lattner2016imposing}, which was developed for the generation of MIDI polyphonic music. In this work, convolution was performed in the time domain to model temporally invariant motives.

% C-RNN-GAN: mogren2016c

%%% NOTE: See 4.4.8 Transposition (augmentation)

Some works have exploited GANs for generating music \cite{mogren2016c, yang2017midinet}. An example of the use of GAN is C-RNN-GAN \cite{mogren2016c} with both $G$ and $D$ being LSTMs in which the goal was to transform random noise into melodies. A bidirectional RNN was utilized in $D$ to take contexts from both past and future. In \cite{yang2017midinet}, a convolutional GAN architecture named MidiNet was proposed to generate pop music melodies from random noise (in piano-roll like format). In this approach, both $G$ and $D$ were composed of convolutional networks. Similar to what a recurrent network would do in considering the history, the information from previous musical measure was incorporated into intermediate layers.

\section{Semi-Recurrent CNN-based VAE-GAN}
\label{Proposed Method: Hybrid VAE-GAN}

% continuous sequential data: temporal data
In this section, the semi-recurrent hybrid VAE-GAN model proposed for generating temporal data such as music, is described. As illustrated in Figure \ref{fig:trainingframework}, the model is composed of three units: the encoder ($E$), the generator/decoder ($G$), and the discriminator ($D$). In this work, the VAE decoder and the GAN generator are collapsed into one by letting them share parameters and training them jointly.

The main architecture of the three networks used in this work is CNNs. Convolutions are rarely used in modelling signals with invariance in time such as music, but they have been very successful in the models whose data has strong spatially local correlation such as images, which is also important for sequential data. In this work, we consider the input time-dependent data as a sequence of individual frames, which have internal spatial correlation. Thus, we exploit CNNs for separate generation of each of these frames, while keeping the dependencies across them as follows. 

For each pair of sequential frames, the previous frame is encoded to its corresponding latent representation using $E$. Next, $G$ tries to generate (predict) the subsequent frame from the latent distribution of the previous frame. As a result, the history and the information from previous frames are incorporated for generating the next ones. The current real training frame in each pair and the synthesized frame are then forwarded to $D$ as real and fake data, respectively.

%CNNs are practically faster to train and more easily to parallelize than RNNs. %

%as opposed to images where motives are a priori invariant in all dimensions

%Wavenet: Although the basis of the architecture is a feed-forward network and not a recurrent network, the transversal causal connexions make it analog to a recurrent network (see section "10.1 Convolution vs Recurrent")

%Wavenet: The authors mention that enlarging the "length of the time window (receptive field)" processed was crucial to obtain samples that sounded musical.

% network architecture

\subsection{Formulation and Objective}
\label{Formulation}

%and  $\{x^{t-1},x^t\}$ be one pair of training data at time $t-1$

Let $X=\{x^0,...,x^{t-1},x^t,...,x^n\}$ be a sequence from the training data with $n$ frames, the network $E$ maps a training frame $x^{t-1}$ (previous time frame) to the mean $\mu$ and the covariance $\epsilon$ of the latent vector:
\begin{equation}
\{\mu,\epsilon\} = E(x^{t-1}) = q(z|x^{t-1}).
\label{eq:mean and covariance}
\end{equation}
Then, the latent vector $z^t$ can be sampled as follows:
\begin{equation}
z^t = \mu + z^t_p \odot exp(\epsilon),
\label{eq:latent vector}
\end{equation}
where $z^t_p \sim \mathcal{N}(0, I)$ and $\odot$ is the element-wise multiplication. In order to reduce the gap between the prior $p(z^t)$ and the encoder's distribution $q(z|x^{t-1})$ and measure how much information is lost, KL loss is used:
\begin{equation}
\mathcal{L}_{prior}=\mathcal{L}_{KL}=\frac{1}{2}(\mu^T\mu+sum(exp(\epsilon)-\epsilon-1)).
\end{equation}

The network $G$ then generates two frames $\tilde{x}^t$ and $\tilde{x}^t_p$ 
%by  sampling from learned distributions $p(x|z^t)$ and $p(x|z^t_p)$, respectively:
by decoding the latent representations $z^t$ (sampled using $E$) and $z^t_p$ (sampled from a normal distribution) back to the data space, respectively:
\begin{equation}
\tilde{x}^t=G(z^t), \hspace{5pt} \tilde{x}^t_p=G(z^t_p).
\end{equation}

Element-wise reconstruction errors are generally inadequate for signals with invariances \cite{larsen2015autoencoding}. As a result, in order to measure the quality of the reconstructed samples in this work, the following pair-wise feature matching loss between the real data $x^t$ and the synthesized data $\tilde{x}^{t}$ and $\tilde{x}^t_p$ is utilized:
\begin{equation}
\mathcal{L}_{l} = \frac{1}{2}{\lVert D_l(x^t)-D_l(\tilde{x}^{t}) \rVert}^2_2 + \frac{1}{2}{\lVert D_l(x^t)-D_l(\tilde{x}^t_p) \rVert}^2_2,
\label{eq:pair-wise feature matching}
\end{equation}
where $D_l$ denotes the features (hidden representation) of an intermediate layer of the network $D$. Thus, the loss of network $E$ is calculated: 
\begin{equation}
\mathcal{L}_{E} = \mathcal{L}_{l} + \mathcal{L}_{prior}.
\end{equation}

In order to distinguish the real training data $x^t$ from the synthesized frames $\tilde{x}^{t}$ and $\tilde{x}^t_p$, the following objection function is minimized by $D$:
% or minimize with multiplying a minue!
\begin{equation}
\mathcal{L}_D = -(\log D(x^t) + \log(1 - D(\tilde{x}^{t})) + \log(1 - D(\tilde{x}^t_p))),
\end{equation}
while $G$ tries to fool $D$ by minimizing
% or maximize without a minus sign!
\begin{equation}
\mathcal{L}_G = - (\log D(\tilde{x}^{t}) + \log D(\tilde{x}^t_p)) + \mathcal{L}_{l},
\end{equation}
where $\mathcal{L}_{l}$ is the pair-wise feature matching loss (Equation \ref{eq:pair-wise feature matching}), which is a shared error signal between $E$ and $G$.

Finally, our goal is to minimize the following hybrid loss function: $\mathcal{L} = \mathcal{L}_{E} + \mathcal{L}_{D} + \mathcal{L}_{G}$.

% or maximize!
%\begin{equation}
%\mathcal{L}_D = \log D(x^t) + \log(1 - D(\tilde{x}^{t})) + \log(1 - D(\tilde{x}^t_p))
%\end{equation}

%The network G tries to learn the real data distribution by the gradients given by the discriminative network D which learns to distinguish between “real” and “fake” samples

%Similar to VAE, for each sample, the encoder network outputs the mean and covariance of the latent vector,i.e., µ and $\epsilon$.

%- Explain the whole framework (based on the figure)\\

%one pair of bars (t, t-1) is given to the encoder, encoder gives the mean and covariance, z generated from mean and covariance, \\

% objective functions:

%\begin{figure}[htb]
%\begin{minipage}[b]{1.0\linewidth}
%  \centering
%  \centerline{\includegraphics[width=8.5cm]{localized.png}}
%  %\centerline{Subtraction}\medskip
%\end{minipage}
%\caption{...}
%\end{figure}

\section{Experiments: Piano Music Generation}
\label{Experiments}

We applied the proposed approach to piano music generation. The source code and some generated samples are shared on GitHub\footnote{https://github.com/makbari7/SR-CNN-VAE-GAN}.
%The method proposed in this work was evaluated on a PC with an Intel Core i7-4470 CPU (3.40 GHz) and 8.00 GB RAM. All the videos and images used in this evaluation were captured using an HD 720p webcam with resolution and frame rate of 640$\times$360 pixels and 30 FPS. A stable stand was used to hold the camera at the top of the piano keyboard to capture a video of the pianist’s performance (similar to \cite{myieeepaper}).
% Howard: if there is no space to include figure, we can just say "the camera is located at the top of the keyboard in the way as \cite{myieeepaper}" or something like that.
%\subsection{Setup}
%\label{Setup}
%which dataset has been used? Nottingham\\
%We tested our proposed model on piano music generation.
In this experiment, we used the Nottingham dataset \footnote{http://www.iro.umontreal.ca/~lisa/deep/data} as our training data, which contains 695 pieces of folk piano music in MIDI file format. Each MIDI file was divided into separate bars, and a bar is represented by a real-valued 2-D matrix $x \in [0, 1]^{h\times w}$ where $h$ and $w$ represent the number of MIDI notes/pitches (i.e., $h=88$ in this work) and the number of time steps (i.e., $w=16$ with pitch sampling of 0.125sec), respectively. The value of each element of the matrix is the velocity (volume) of a note at a certain time step. The sequence of $n$ bars is denoted by $X=\{x^0,...,x^{t-1},x^t,...,x^n\}$ where $x^{t-1}$ and $x^t$ are two sequential bars. 

%employ convolutions on a 2-D matrix representing the presence of notes over different time steps in a bar. We can have such a score-like representation for each bar for either a real or a generated MIDI.

%This way, our model can “look back” without a recurrent unit as used in RNNs. Like RNNs, our model can generate music of arbitrary number of bars

%Network achitecture:
%E: conv (8), conv (16), conv (32), FC (2) {5*5, s=2}
%D: 16, 32, 64, linear, sigmoid {3*3, s=1}
%G: deconv: linear, 16, 32, 64 {3*3, s=1} 

%The same generator and discriminator networks as DCGAN were used

The details of the networks $E$, $G$, and $D$ are summarized in Table \ref{tbl:network}. The output layer of $E$ is a fully-connected layer with 256 hidden units where its first and second 128 units are respectively considered as the mean $\mu$ and covariance $\epsilon$ used for representing the latent $z^t$ of dimension 128 (Equations \ref{eq:mean and covariance} and \ref{eq:latent vector}). The latent $z^t$ and a normal distribution $z^t_p$ (of dimension 128) are projected to $G$ to output the synthesized bars $\tilde{x}^t, \tilde{x}^t_p \in [0, 1]^{88\times 16}$. Before the Tanh layer of $G$, another convolution is applied to map to the number of output channels (that is 1 in this work). An extra convolution is also applied before the Sigmoid layer of $D$ to represent the output by a 1-D feature map, which is used as $D_l$ for calculating the pair-wise feature matching loss (Equation \ref{eq:pair-wise feature matching}). This network takes the 2-D matrices $x^t$ and $\tilde{x}^t$ as inputs and predicts whether they are real or generated MIDI bars.

\begin{table}[htb]
\scriptsize
\centering
\caption{The network architecture of the encoder (\textbf{E}), generator (\textbf{G}), and discriminator (\textbf{D}). \textbf{AF}, \textbf{In}, and \textbf{Out} are respectively the activation functions used after each conv/deconv layer, the input, and the output of each network.
}
\label{my-label}
\begin{tabular}{c|c|c|c|c|c|}
\cline{2-6}
& \textbf{Layers (filters)}            & \textbf{Size} & \textbf{AF} & \textbf{In}      & \textbf{Out}               \\ \hline
\multicolumn{1}{|c|}{\textbf{E}} & \begin{tabular}[c]{@{}c@{}}conv (8, 16, 32), \\ Fully-connected layer\end{tabular}   & \begin{tabular}[c]{@{}c@{}}5$\times$5\\ stride=2\end{tabular} & ELU                   & $x^{t-1}$          & \{$\mu, \epsilon$\}              \\ \hline
\multicolumn{1}{|c|}{\textbf{G}} & \begin{tabular}[c]{@{}c@{}}deconv (64, 32, 16, 8), \\ Tanh layer\end{tabular} & \begin{tabular}[c]{@{}c@{}}3$\times$3\\ stride=1\end{tabular} & ReLU                  & $z^t, z^t_p$       & $\tilde{x}^t, \tilde{x}^t_p$ \\ \hline
\multicolumn{1}{|c|}{\textbf{D}} & \begin{tabular}[c]{@{}c@{}}conv (8, 16, 32, 64), \\ Sigmoid layer\end{tabular}   & \begin{tabular}[c]{@{}c@{}}3$\times$3\\ stride=1\end{tabular} & LeakyReLU             & $x^t, \tilde{x}^t$ & 0 or 1                       \\ \hline
\end{tabular}
\label{tbl:network}
\end{table}

%The activation functions used in $E$, $G$, and $D$ are ELU (exponential linear unit) \cite{clevert2015fast}, ReLU (rectified linear unit), and LeakyReLU (leaky rectified linear unit), respectively. Inspired by DCGAN, batch normalization was applied to all layers in all three models and no pooling layer was used. 
All models were trained with mini-batch stochastic gradient descent (SGD) with a mini-batch size of 64. The Adam optimizer with momentum of 0.5 and learning rate of 0.0005 for $E$ and $G$, and 0.0001 for $D$ was used. In order to keep the losses corresponding to $E$, $G$, and $D$ balanced in each iteration, we trained $E$ and $G$ twice and $D$ once.

\begin{figure}[h!]
\begin{minipage}[b]{1.0\linewidth}
 \centering
  \centerline{\includegraphics[width=8.6cm]{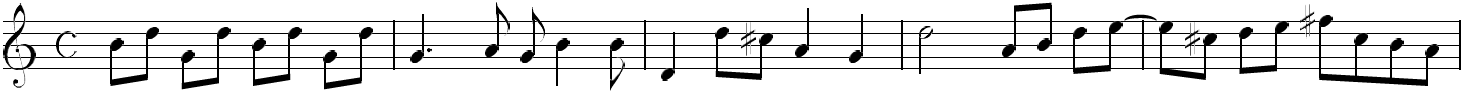}}
  %\centerline{Subtraction}\medskip
\end{minipage}
%\\
%\\
%\begin{minipage}[b]{1.0\linewidth}
%  \includegraphics[width=8cm]{280.png}
%  %\centerline{Subtraction}\medskip
%\end{minipage}
\\
\\
\begin{minipage}[b]{1.0\linewidth}
  \includegraphics[width=7cm]{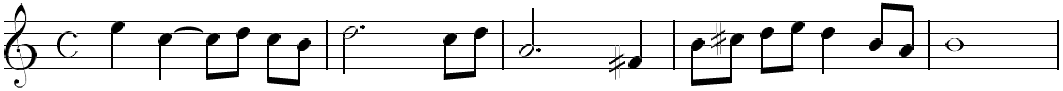}
  %\centerline{Subtraction}\medskip
\end{minipage}
\caption{Two 5-bar sample music generated using the proposed testing model 1 (top model in Figure \ref{fig:testingframework}).}
\label{fig:musicscore}
\end{figure}

Two models illustrated in Figure \ref{fig:testingframework} were proposed to sequentially generate music with an arbitrary number of bars. In model 1 (top model in Figure \ref{fig:testingframework}), the input to $E$, denoted by $x^0$, is a bar randomly selected from training data samples, which is considered as the first bar of the generated music. $x^0$ is then mapped to the latent $z^1$ using $E$. $G$ synthesizes the next bar $\tilde{x}^1$ by decoding $z^1$ back to the data space. By feeding the generated bar $\tilde{x}^1$ to $E$, this process is repeatedly performed to generate a sequence of bars. In model 2 (bottom model in Figure \ref{fig:testingframework}), the same recurrent process is applied, but the first bar is also a bar synthesized using $G$ from a random noise $z_p$. Two 5-bar sample music generated using model 1 (top model in Figure \ref{fig:testingframework}) are illustrated in Figure \ref{fig:musicscore}.

\subsection{Results}
\label{Results}

\begin{figure}[h!]
\begin{minipage}[b]{1.0\linewidth}
 \centering
  \centerline{\includegraphics[trim={0.5cm 4.5cm 0.5cm 4cm},width=9.5cm]{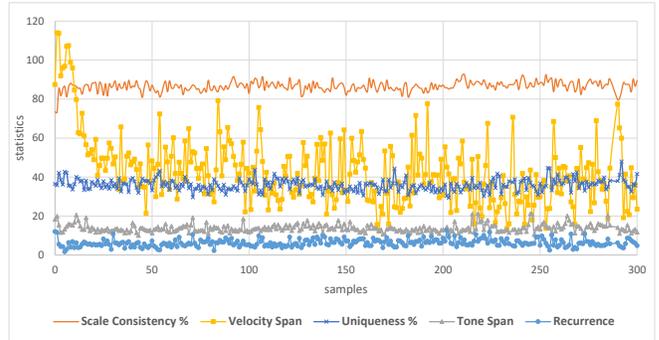}}
  %\centerline{Subtraction}\medskip
\end{minipage}
\caption{Measurements used for evaluating $\approx{2,500}$ music samples generated at 300 epochs: scale consistency, intensity span, uniqueness, tone span, and recurrence.}
\label{fig:statistics}
\end{figure}

In order to evaluate the music samples generated using our approach, the following measurements were taken into account \cite{mogren2016c}: \textbf{scale consistency} (the percentage for the best matching musical scale that a sample is part of), \textbf{uniqueness} (the percentage of unique tones used in a sample), \textbf{velocity span} (the velocity range in which the tones are played), \textbf{recurrence} (repetitions of short subsequences of length 2 in a sample), \textbf{tone span} (the number of half-tone steps between the lowest and the highest tones in a sample), and \textbf{diversity} (the average pairwise Levenshtein edit distance \cite{habrard2008melody} of the generated data ). Figure \ref{fig:statistics} shows the results of evaluating $\approx2,500$ generated pieces of music of length 10 seconds (i.e., 5 two-second bars).

As seen in Figure \ref{fig:statistics}, the scale consistency (with an average of $\approx87\%$) shows that the generated music significantly follows the standard scales in all samples, which outperforms C-RNN-GAN \cite{mogren2016c} with an average of $\approx75\%$.
%except the first few ones. 
A variety of velocities exist in the music generated, which is illustrated by the oscillating velocity span. The average percentage of the unique tones used in the generated piece is $\approx37\%$. Compared to the velocity span, less variability is seen in the tone span (with minimum and maximum of 10 and 21) of the generated music due to the low tone span in the training samples (the majority of the music in the dataset is played in 1 or 2 octaves). The number of 2-tone repetitions is $\approx7$ in average. Diversity is another metric we took into account to evaluate how realistic the generated music sounds. Compared to ORGAN \cite{guimaraes2017objective} with an average of 0.551, a higher diversity with an average of $\approx0.59$ was achieved in this work.

% show some diagrams, figures, sample sheet music or piano-rolls, comparison

\section{Conclusion}
\label{Conclusion}
A semi-recurrent VAE-GAN model for generating sequential data was presented in this work. The model consisted of three networks (encoder, generator, and discriminator) in which convolutions were utilized to spatially learn the local correlation of the data in individual frames. Each frame was sampled from a latent distribution obtained by mapping the previous frame using the encoder. As a consequence, the consistencies between the frames in a generated sequence was also preserved. Our experiments on piano music generation presented promising results, which were comparable to the state-of-the-art. One potential direction of this work is to use this framework for modelling and generating other types of sequential data such as video.

\section{Acknowledgement}
This work was supported by the Natural Sciences and Engineering Research Council (NSERC) of Canada under grant RGPIN312262 and RGPAS478109.

\newpage
 % Revious frameferences should be produced using the bibtex program from suitable
%\bibliographystyle{IEEEbib}

\end{document}